%% file: main.tex
\documentclass{article} % For LaTeX2e

\usepackage{iclr2023_conference,times}

\input{math_commands.tex}

\input{macro}

\usepackage{hyperref}
\usepackage{url}
\usepackage{graphicx}

\usepackage{ragged2e}
\usepackage{tabularx}
\title{RETSim: Resilient and Efficient Text \newline Similarity}

\author{Marina Zhang\textsuperscript{1}, Owen Vallis\textsuperscript{1}, Aysegul Bumin\textsuperscript{*2}, Tanay Vakharia\textsuperscript{1}, Elie Bursztein\textsuperscript{1} \\
Google\textsuperscript{1} \quad  University of Florida\textsuperscript{2}\\
}

\iclrfinalcopy % Uncomment for camera-ready version, but NOT for submission.
\begin{document}

\maketitle

\makeatletter\def\Hy@Warning#1{}\makeatother %https://tex.stackexchange.com/questions/415625/avoiding-hyperref-warning-ignoring-empty-anchor
\def\thefootnote{*}\footnotetext{This work was done during the author’s internship at Google.}\def\thefootnote{\arabic{footnote}}

\input{abstract}

\input{introduction}

\input{related}

\input{restim_architecture}

\input{evaluation}

\input{applications}

\input{ablation}

\input{conclusion}

\bibliography{refs}
\bibliographystyle{iclr2023_conference}

\newpage
\appendix
\input{appendix}
% \section{Appendix}

\end{document}

%% file: math_commands.tex
\usepackage{amsmath,amsfonts,bm}

\def\eqref#1{equation~\ref{#1}}
\def\1{\bm{1}}

\DeclareMathAlphabet{\mathsfit}{\encodingdefault}{\sfdefault}{m}{sl}
\SetMathAlphabet{\mathsfit}{bold}{\encodingdefault}{\sfdefault}{bx}{n}

%% file: macro.tex
\usepackage{xspace}

\newcommand\rs{RETSim\xspace}
\newcommand\rv{RETVec\xspace}

\newcommand\rpd{RETSim\textsubscript{Partial-Dup}\xspace}
\newcommand\rnd{RETSim\textsubscript{Near-Dup}\xspace}
\newcommand\wa{W4NT3D\xspace}

\usepackage{tabularx}
\usepackage{float}
\usepackage{booktabs}
\usepackage{ulem}
\usepackage{multirow}
\usepackage{capt-of}% or \usepackage{caption}
\usepackage{lscape}

%% file: abstract.tex
\begin{abstract}

This paper introduces \rs (Resilient and Efficient Text Similarity), a lightweight, multilingual deep learning model trained to produce robust metric embeddings for near-duplicate text retrieval, clustering, and dataset deduplication tasks. 
We demonstrate that \rs is significantly more robust and accurate than MinHash and neural text embeddings, achieving new state-of-the-art performance on dataset deduplication, adversarial text retrieval benchmarks, and spam clustering tasks. We also introduce the \wa benchmark (Wiki-40B 4dversarial Near-T3xt Dataset) for evaluating multilingual, near-duplicate text retrieval capabilities under adversarial settings. \rs and the \wa benchmark are open-sourced under the MIT License at https://github.com/google/unisim.
\end{abstract}

%% file: introduction.tex
\section{Introduction}
\label{sec:int}

Robust near-duplicate text detection is an essential component of many tasks, including retrieving documents, detecting plagiarism~\citep{hutchison_near_2013} and blocking adversarial spam campaigns~\citep{ahmed_machine_2022}. Users have come to expect that systems can return accurate results despite their queries exhibiting a 20\% to 30\% typo rate ~\citep{hagen_large-scale_2017}. Furthermore, efficiently deduplicating text datasets is critical to training state-of-the-art large language models~\citep{lee_deduplicating_2022,kandpal_deduplicating_2022}.

For more than two decades, MinHash-based ~\citep{broder_min-wise_1998} locality-sensitive hashing (LSH) has been the most prevalent algorithm used for near-duplicate detection due to its simplicity, robustness, and speed. For example, the vast majority of dataset deduplication efforts still rely on MinHash ~\citep{lee_deduplicating_2022,kocetkov_stack_2022}. However, like all LSH-based techniques, MinHash is not without downsides; chief among them being that it is very parameter-sensitive and requires heavy tuning. Additionally, MinHash lacks resilience to typos due to its reliance on n-grams, leading to poor performance on noisy data and a vulnerability to hash-busting attacks~\citep{issac2014analysis}.

On the other hand, deep learning models are the dominant way to perform vector-based semantic text retrieval~\citep{muennighoff_mteb_2022}, but so far, no neural embedding has been able to consistently outperform MinHash for robust near-duplicate detection~\citep{silcock_noise-robust_2022}. This is mostly due to the focus on improving semantic capabilities, which leads models to be too large to run extremely quickly and the use of sub-word level tokenization, which is not resilient to typos and adversarial attacks ~\citep{morris_textattack_2020,bursztein_retvec_2023}.

To fill this gap, we introduce \rs (Resilient and Efficient Text Similarity), a lightweight, multilingual deep learning model trained specifically to produce robust neural embeddings specialized for near-duplicate detection.
By combining the state-of-the-art RETVec text vectorizer, a modern transformer block~\citep{hua_transformer_2022}, a large typo-augmented training corpus, and a metric learning training regime, \rs is able to achieve new state-of-the-art performance on near-duplicate detection benchmarks (Section~\ref{sec:wanted}), dataset deduplication tasks (Sections~\ref{sec:dedup} and \ref{sec:dataset_deup}), and spam clustering applications (Section~\ref{sec:spam_email}).

Furthermore, while datasets and benchmarks exist for corpus deduplication and near-duplicate text retrieval, none of these have focused on systematically evaluating near-duplicate retrieval performance under the presence of typos, word manipulations, and sentence or paragraph-level modifications. To address this need, we additionally introduce the {\it W4NT3D} benchmark (Wiki-40B 4dversarial Near-T3xt Dataset) which enables the evaluation of algorithms on adversarial near-duplicate text retrieval in a multilingual setting. We report the performance of \rs, MinHash, and popular neural embeddings such as Universal Sentence Encoder~\citep{cer_universal_2018} and LaBSE~\citep{feng_language-agnostic_2022} on this new benchmark in Section~\ref{sec:wanted}, highlighting uneven performance across languages and types of adversarial manipulations. The RETSim model and the W4NT3D benchmark are open-sourced at https://github.com/google/unisim under the MIT License.

%W4NT3D results demonstrate \rs's superior effectiveness in adversarial settings while also exploring its limitations.

% \item We propose \rs,  

%% file: related.tex
\section{Related Work}
\label{sec:rel}
% \fixme{marina 1 page}

% In this section we review related work in the area of syntactic text similarity, including applications of text matching, text retrieval, edit distance, locality sensitive hashing (LSH), semantic text similarity, and approximate nearest neighbor methods. Additionally, we will also discuss the impact of adversarial text attacks on these similarity measures.

\paragraph{Near-Duplicate Detection}
Identifying noisy near-duplicate documents in a large corpus is a fundamental task with a wide range of applications, such as detecting plagiarism, finding reproduced content in literature or news articles~\citep{gyawali_deduplication_2020, silcock_noise-robust_2022}, and deduplicating training datasets for language models. Previous research has shown that duplicates in training datasets lead to inefficient training~\citep{lee_deduplicating_2022} and privacy concerns for large language models (LLMs), where models memorize and regenerate duplicated training sequences at a much higher frequency~\citep{kandpal_deduplicating_2022}.

Unlike semantic text similarity, the task of identifying textual near-duplicates has been predominated by non-neural, n-gram-based algorithms such as MinHash~\citep{broder_min-wise_1998}, which is the most widely used technique for deduplicating large training corpuses~\citep{kocetkov_stack_2022, lee_deduplicating_2022}. MinHash  is a technique for estimating the Jaccard similarity between two sets. Algorithms such as MinHash or SimHash~\citep{charikar_similarity_2002} can be combined with locality-sensitive hashing (LSH) techniques for fast, approximate nearest neighbor search and data clustering. This allows them to scale and deduplicate corpuses containing terabytes of data such as C4~\citep{lee_deduplicating_2022} and The Stack~\citep{kocetkov_stack_2022}. However, n-gram or shingling-based techniques typically require texts to be parsed into a standardized form (e.g. by lower-casing or stripping punctuation), which makes them susceptible to typos and adversarial attacks and pose a challenge when attempting to differentiate between dissimilar documents and near-duplicate documents with adversarial augmentations.

\paragraph{Semantic Text Similarity}

The task of computing semantic similarity between text is closely related to near-duplicate detection. Semantic text similarity refers to the assessment of the semantic relatedness of two pieces of text based on their meaning rather than their syntactic structure, as in the case of near-duplicate detection. Recently, transformer-based language models such as Universal Sentence Encoder~\citep{yang_multilingual_2019}, LaBSE~\citep{feng_language-agnostic_2022} and LLM-based embeddings~\citep{anil_palm_2023} which embed text into high-dimensional embedding vectors have been successfully used to retrieve semantically-related documents using cosine similarity. Modern text retrieval systems combine these embeddings with an approximate nearest neighbor (ANN) search algorithm to efficiently retrieve documents matching user queries.

However, language models have been shown to be vulnerable to adversarial attacks and naturally-occurring typos~\citep{alzantot_generating_2018,gao_black-box_2018, morris_textattack_2020}. Furthermore, language models are typically very large and costly to run even with hardware acceleration, which makes them unsuited for large-scale dataset deduplication or identifying near-duplicates in the presence of typos or adversarial text manipulations.

\paragraph{Metric Learning} 
Metric learning aims to learn an embedding space where similar items have a small distance between their embeddings and dissimilar items are further away. Many state-of-the-art embeddings use metric learning for unsupervised training or fine-tuning including Sentence-BERT~\citep{reimers_sentence-bert_2019} and E5~\citep{wang_text_2022}.

\rv is a resilient, multilingual embedding and text vectorizer trained to be robust against various forms of character-level typos and adversarial attacks. We extend the \rv training regime to full text documents for \rs. We use Multi-Similarity Loss~\citep{wang_multi-similarity_2019} for pair-based metric learning, where typo-laden and near-duplicate versions of texts are trained to be closer in the embedding space, while other texts are pushed further away. Multi-Similarity Loss is based on a general weighting framework for pair-based losses and achieves state-of-the-art performance, outperforming alternatives such as Triplet Loss~\citep{schroff_facenet_2015}.

%% file: restim_architecture.tex
\section{RETSim}
\label{sec:arc}
\begin{figure}[t]
\vspace{-1mm}
\centering
\scalebox{1.0}{
\includegraphics[width=\textwidth]{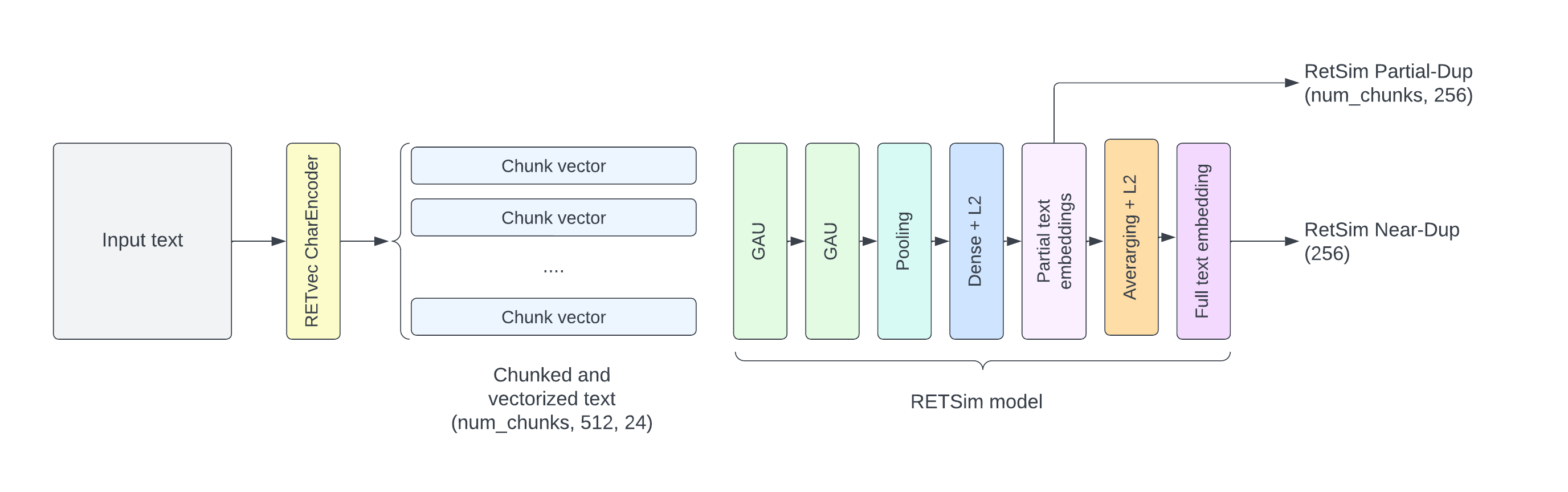}}
\vspace{-8mm}
\caption{\rs model architecture diagram. \rs works on arbitrary length text by splitting texts into chunks of 512 characters during its vectorization phase and encodes them using the RETVec character vectorizer. The \rs model then embeds each chunk of text into 256-dim partial embeddings and combines them to produce the global embedding.}
\label{fig:arc}
\end{figure}

% In this section, we present an overview of \rs. We start by discussing the \rs model architecture, then we describe the dataset used to train \rs, and finally we summarize the \rs training regime.

\subsection{Architecture}
The \rs model is composed of three main components (as depicted in Figure~\ref{fig:arc}):

\paragraph{The character-level vectorizer} splits the input text into chunks of 512 characters, then uses the RETVec chararcter encoder~\citep{bursztein_retvec_2023} to encode each chunk, resulting in a batch of $(512, 24)$ dense inputs. The RETVec character vectorizer encodes each Unicode character as a compact 24-bit binary representation based on its integer codepoint value. This allows the vectorizer to encode all valid Unicode characters and support all languages. Furthermore, the character-level vectorizer has been shown to be more resilient against typos and adversarial attacks.

\paragraph{A small transformer model} is used to compute 256-dimension embeddings for each chunk of the input text. \rpd uses these embeddings directly to finding documents that have matching chunks of text. Architecturally, the model consists of two Gated Attention Unit (GAU) blocks~\citep{hua_transformer_2022}, followed by a Generalized-Mean pooling layer ~\citep{radenovic_fine-tuning_2018}, a dense projection layer which projects the embedding into 256 dimensions, and an L2 normalization layer. The model has only 536k parameters, which is more than two orders of magnitude smaller than other neural embeddings (Table~\ref{tab:algos}). L2-normalization allows the embeddings to be compared using cosine similarity. We discuss the impact of key architecture design choices in Section~\ref{sec:abl}. Hyperparameter details are provided in Appendix~\ref{sec:app:mdl:hyp}, and additional ablations results in Appendix~\ref{app:ablation}.

\paragraph{An embedding averaging module} is then used to combine partial text embeddings into a full-text embedding which is used for global near-duplicate matching (\rnd). Averaging chunked embeddings to produce a global embedding is a standard technique used by many models~\citep{cer_universal_2018} to support infinite length inputs in a cost-efficient manner. We experimented with other aggregation techniques to produce more accurate global embeddings, including training a deep-averaging network \citep{iyyer_deep_2015}, but this did not improve performance and resulted in higher computation cost. \rnd and \rpd are computed in a single forward pass which makes it computationally efficient. We output both types of embeddings as they have different applications: \rnd is better-suited for full-text matching and retrieval (Section~\ref{sec:wanted}), while \rpd is used to find partial text matches where the near-duplicate content appears only in part of the document (Section~\ref{sec:dedup}).

% is used to find partial text matches. 

% Both are needed because past a certain text length the full text embedding loose the ability to accurately represent the finer details of each chunk as reported in~\ref{sec:dedup}

% Both 
% It is also important to note that the embeddings for \rnd and \rpd are computed in a single forward pass which makes it computationally efficient. In Section~\ref{sec:wanted}, we show that for adversarially robust text retrieval on full documents longer than 512 characters, \rnd outperforms \rpd since the averaged embedding represents the entire text. On the other hand, for near-duplication detection use-cases where the near-duplicate content appears only in part of the document (Section~\ref{sec:neardup}), \rpd is able to effectively find parts of documents that are syntactically similar.

\subsection{Model Training}
\label{sec:training_dataset}

\paragraph{Dataset}  We use the multilingual C4 dataset (mC4) for raw text data and following ~\citep{xue_mt5_2020}, we use a language sampling exponent of $\alpha=0.3$ to balance sampling between low and high-resource languages. We only use text containing at least 16 characters, and we randomly select between 1 and 8 sentences (roughly 512 characters) for each text chunk. For each example in the training dataset, we generate 5 pairs of augmented examples. We apply three levels of augmentation to each example text chunk (in this order): sentence-level, word-level, and character-level. For each level, we randomly select the augmentation to be applied from the following categories: insertion, deletion, substitution, and transposition. We randomly apply between $0-25\%$ sentence-level augmentation and up to $30\%$ combined character and word-level augmentation. Empirically, we found that increasing the percentage of augmentation beyond this point causes \rs's performance to degrade. The full list of augmentations used can be found in Appendix~\ref{sec:app:aug}.

\paragraph{Training Procedure} 
We train \rs using Multi-Similarity Loss~\citep{wang_multi-similarity_2019} with $\alpha=4$, $\beta=40$, $\lambda=0.5$, and $\epsilon=0.1$. We hypertuned these parameters and the results are shown in Appendix~\ref{app:ablation}. We train for 1 million steps with batch size $=1024$. The similarity loss trains the model to embed augmented versions of the same text closer in the embedding space, while dissimilar texts are pushed further apart. We use the LAMB optimizer~\citep{you_large_2019} with a max learning rate of $0.001$ and cosine decay. Detailed training hyperparameters are reported in Appendix~\ref{sec:app:tra}.

% \subsection{RETSIM\textsubscript{Near-Dup} and RETSIM\textsubscript{Partial-Dup}} 

% \fixme{Help please at frameing retsim partial and near dup? }

% Both methods allow \rs to accept inputs of arbitrary length. It is also important to note that the embeddings for \rnd and \rpd are computed in a single forward pass which makes it computationally efficient to compute both \rnd similarity and \rpd similarity values. As we show in the evaluations, \rnd and \rpd both have their use-cases. In Section~\ref{sec:wanted}, we show that for adversarially robust text retrieval on full documents longer than 512 characters, \rnd outperforms \rpd since the averaged embedding is able to represent and match based on similarity of the entire text. On the other hand, as shown Section~\ref{sec:neardup}, in cases of near-duplication detection where the near-duplicate content appears only in part of the document, \rpd is able to find parts of documents that are syntactically simiar near-duplicates of each other.

% \note{TODO: Add discussion looking at quantize the dtype as well as the number of dims. Input size is 512*8 and the output is 256*32}

%% file: evaluation.tex
\section{Evaluation}

\label{sec:eval}
% In this section, we evaluate \rs's performance against state-of-the-art multilingual, semantic text embeddings such as E5~\citep{wang_text_2022} as well as hash-based near-duplicate detection algorithms like MinHash~\citep{broder_min-wise_1998} on three different benchmarks. For the first benchmark, we create \wa, a new, multilingual benchmark based on the Wiki-40B dataset which evaluates an algorithm's ability to retrieve near-duplicate text in the presence of typos or other textual modifications. We also benchmark \rs and other baseline algorithms on the NEWS-COPY Deduplication dataset~\citep{silcock_noise-robust_2022} and on the CORE Near-Duplicates dataset~\citep{gyawali_deduplication_2020}.

% introduced in the  
% near-duplication detection  
% The second and third benchmarks are near-duplicate text detection datasets 
% For the second and third benchmarks,
% we benchmark \rs and other baseline algorithms on the NEWS-COPY de-

% against classic near-duplicate detection algorithms such as MinHash and state-of-the-art neural embeddings including Universal Sentence Encoder, LaBSE, and E5 on three benchmarks. The first benchmark, \wa, \fixme{say we create this one and say it's novel}

% evaluates an algorithm's ability to retrieve adversarially augmented text.
% The second and third benchmarks, which are classically used in the literature (NEWS-COPY~\cite{} and CORE Near-Duplicate Dataset ~\cite{}), evaluate algorithms on their ability to detect near-duplicates on real world datasets.
% \vspace{-2mm}

\begin{table}[h]
\centering
\scalebox{0.8}{\begin{tabular}{@{}l|ccc@{}}
\toprule
\textbf{Model/Algorithm} & \textbf{Type} & \textbf{Embed./Hash Size} & \textbf{\# Model Parameters} \\ \midrule
LaBSE & Neural & 768 & 471M \\
Multilingual USE & Neural & 512 & 69M \\
Multilingual E5-Base & Neural & 768 & 278M \\ 
PaLM 2 (Gecko) & Neural & 768 & ?  \\\midrule
SimHash & Hashing & \textit{b} bits & N/A \\
MinHash & Hashing & \textit{n} hashes & N/A \\  \midrule
RETSim & Neural & 256 & 536k \\ \bottomrule
\end{tabular}}
\caption{Embedding models and hashing algorithms benchmarked in the paper.}
\label{tab:algos}
\end{table}
\vspace{-1mm}
\subsection{Models and Algorithms Evaluated}

We benchmark \rs against four multilingual semantic text embeddings as well as popular n-gram based algorithms primarily used in near-duplicate text detection (Table~\ref{tab:algos}). Our baseline text embeddings include Multilingual Universal Sentence Encoder ~\citep{yang_multilingual_2019}, LaBSE~\citep{feng_language-agnostic_2022}, Multilingual E5~\citep{wang_text_2022}, and PaLM 2 Gecko Embeddings~\citep{anil_palm_2023}. All text embeddings are L2-normalized and compared using cosine similarity. We use exact search to index and retrieve nearest neighbors from our vector index for the experiments in Section~\ref{sec:eval}.
% cosine similarity threshold for labeling a pair of texts as near-duplicates of each other from the set $t=\{0.80, 0.81, 0.82, ..., 0.98, 0.99\}$.

For non-neural near-duplicate detection and clustering algorithms, we selected the two most popular algorithms: MinHash~\citep{broder_min-wise_1998} and SimHash~\citep{charikar_similarity_2002}. For MinHash, we use Datasketch’s MinHashLSH library. Following the most common practices in the literature~\citep{silcock_noise-robust_2022}, we use 10 hash functions for MinHash unless otherwise specified. We use word-level n-grams where we select the best value out of $n=\{2,3,4,...,10\}$. For SimHash, we use 64-bit SimHash and conduct shingling at the character level, where the shingle size is selected from $n=\{2,3,4,...,10\}$.
For the near-duplicate detection benchmarks (NEWS-COPY and CORE Near-Duplicates datasets), we tune the optimal deduplication threshold (e.g. based on cosine similarity for neural-based embeddings and Jaccard similarity for MinHash). Detailed hyperparameter settings for \rs and baseline algorithms used in the evaluation can be found in Appendix~\ref{app:eval_hyp}.
% and the Hamming distance deduplication threshold is selected from $t=\{1,2,3,4,5,6,7,8,9,10\}$.

% \fixme{we can just say we optimize the deduplication threshold and report the results in the appendix if we need space}

% Throughout this section, we are comparing \rs against state-of-the-art semantic text embeddings and near-duplicate detection algorithms reported in Table~\ref{tab:algos}. 
% Two decades in,  with MinHash + LSH remains the most popular algorithms used for dataset deduplication used e.g. the stack, c4, ... 

% More recently, semantic neural embeddings have been used as an alternative to (semantic text retrieval emphasis). In this section, we benchmark the most efficient ones including

\subsection{\wa: Wiki-40B 4dversarial Near-T3xt Dataset Evaluation}
\label{sec:wanted}

% We introduce a multilingual benchmark created from Wiki-40B designed to measure adversarial robustness for syntactic text retrieval. We benchmark the performance of \rs against state-of-the-art semantic text embeddings as well as MinHash and SimHash. \fixme{we need to provide justification for the syntactic text retrieval task
% }

\paragraph{Dataset Description}
The vast majority of text retrieval benchmarks are focused on evaluating semantic performance. To the best of our knowledge, there is no multilingual benchmark for systematically measuring adversarial robustness for near-duplicate text retrieval. In an attempt to fill in the gap, we create and publish the \wa benchmark (Wiki-40B 4dversarial Near-T3xt Dataset), which contains around 400k pairs of syntactically similar texts to evaluate near-duplicate text retrieval in the presence of various forms of text manipulations and typos.

\wa is based on the Wiki-40B dataset~\citep{guo_wiki-40b_2020}. The dataset is split into query examples and target examples, where query examples are synthetically-modified near-duplicate versions of a target example (e.g. with typos). For each of the 41 language splits in Wiki-40B, we randomly select 10,000 texts. The length of the target string is uniformly selected from between 16 and 8192 characters, in order to test performance on short and long text. To construct the query text corresponding to a target text, we randomly apply up to 25\% word and character augmentations, and up to 25\% sentence and paragraph augmentations. For each augmentation, we uniformly select from the [insert, delete, substitute, and swap] operations. We use Recall$@k$ with $k=1$ as the main metric, following the setup commonly found in semantic text retrieval benchmarks. 

% \subsubsection{Results}
\begin{table}[ht]
\centering
\scalebox{0.65}{\begin{tabular}{@{}l|cccccccccc|c@{}}
\toprule
\multicolumn{1}{l|}{\textbf{Model/Algorithm}}  & \multicolumn{1}{l}{\textbf{Arabic}} & \multicolumn{1}{l}{\textbf{Chinese}} & \multicolumn{1}{l}{\textbf{English}} & \multicolumn{1}{l}{\textbf{German}} & \multicolumn{1}{l}{\textbf{French}} & \multicolumn{1}{l}{\textbf{Spanish}} & \multicolumn{1}{l}{\textbf{Japanese}} & \multicolumn{1}{l}{\textbf{Korean}} & \multicolumn{1}{l}{\textbf{Russian}} & \multicolumn{1}{l}{\textbf{Thai}} & \multicolumn{1}{|l}{\textbf{Avg (41 Langs)}} \\ \midrule
LaBSE & 0.915 & 0.917 & 0.944 & 0.931 & 0.930 & 0.888 & 0.931 & 0.949 & 0.918 & 0.882 & 0.921 \\
Multilingual USE & 0.915 & \textbf{0.986} & 0.958 & 0.942 & 0.938 & 0.903 & \textbf{0.990} & 0.984 & 0.910 & 0.888 & 0.912 \\
Multilingual E5-Base & \underline{0.936} & \underline{0.980} & 0.959 & 0.944 & \underline{0.948} & 0.896 & \underline{0.979} & \underline{0.986} & 0.911 & 0.921 & 0.937 \\
PaLM 2 (Gecko) & 0.497 & 0.623 & \underline{0.961} & 0.932 & 0.934  & 0.911 & 0.578 & 0.701 & 0.851 & 0.571 & 0.823 \\  \midrule
SimHash & 0.558 & 0.276 & 0.591 & 0.561 & 0.519 & 0.513 & 0.465 & 0.593 & 0.554 & 0.669 & 0.550 \\
MinHash & 0.633 & 0.172 & 0.591 & 0.558 & 0.556 & 0.575  & 0.223 & 0.814 & 0.523 & 0.416 & 0.538 \\  \midrule
\rpd & 0.928 & 0.946 & 0.954 & \underline{0.949} & 0.947 & \underline{0.938} & 0.963 & 0.971 & \underline{0.946} & \underline{0.941} & \underline{0.949} \\
\rnd & \textbf{0.971} & 0.971 & \textbf{0.987} & \textbf{0.978} & \textbf{0.983} & \textbf{0.976} & 0.986 & \textbf{0.991} & \textbf{0.970} & \textbf{0.946} & \textbf{0.977} \\ \bottomrule
\end{tabular}}
\caption{
Per-language retrieval performance for various embedding models and algorithms on the \wa benchmark. Results on selected languages are reported alongside the average Recall@1 for all 41 languages. Full results for all languages are reported in Appendix~\ref{app:wiki_full}.}
\label{tab:wiki_main}
\end{table}

\vspace{-2mm}

\paragraph{Multilingual Performance}
Overall, \rnd achieves an average Recall@1 of 0.977 across all 41 languages on the \wa benchmark (Table~\ref{tab:wiki_main}). \rpd is second best with a Recall@1 of 0.949 and Multilingual E5, the best-performing baseline, is third with an average Recall@1 of 0.932. We expect that \rnd outperforms \rpd because the \wa benchmark requires an algorithm to not just find near-duplicates, but to find the most similar text. \rpd optimizes for finding the most similar chunk of text in the corpus, which is not always the most similar text overall. Similarly, we hypothesize that MinHash and SimHash perform poorly on the \wa benchmark due to their lack of ability to distinguish which is the most similar text among the near-duplicates detected, and embedding-based models and cosine similarity offer a more fine-grained measure of similarity.

\rnd outperforms baseline algorithms on all languages except for Chinese and Japanese. For these languages, we theorize that semantic embeddings may have the slight edge in performance because their significantly larger model sizes (more than 100x larger than \rs, as shown in Table~\ref{tab:algos}) allow them to have a better representation on languages with large character sets. Furthermore, the sub-word level tokenizers used in the baseline embeddings often treat each character in Chinese or Japanese as individual tokens, which could offer higher resilience to typos.

\paragraph{Adversarial Resilience}

\vspace{-2mm}
\begin{figure}[ht]
\centering
\includegraphics[width=0.92\textwidth]{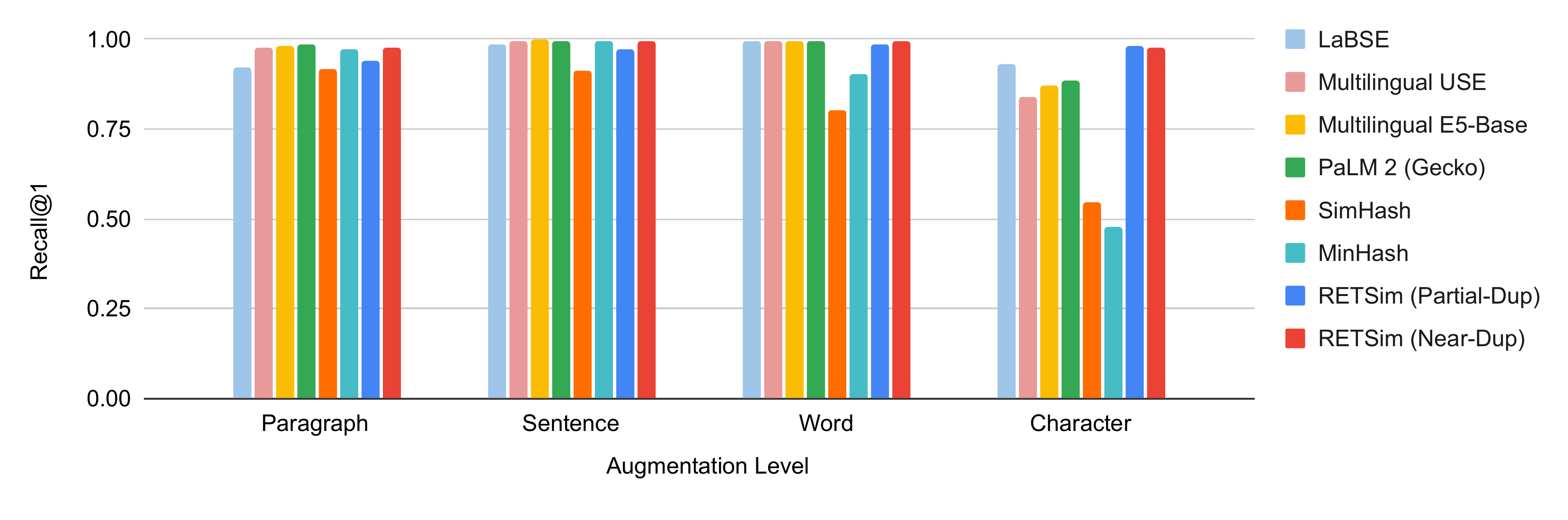}
\vspace{-3mm}
\caption{Recall@1 performance on the \wa benchmark, broken down by augmentation type. Results are averaged across all 41 language splits in \wa.}
\label{fig:aug_level_radar}
\end{figure}
Delving deeper into the impact of various types of text manipulation reveals that \rnd and \rpd perform almost equally well regardless of the type of augmentation applied (Figure~\ref{fig:aug_level_radar}). Semantic text embeddings perform well on paragraph, sentence and word-level manipulations, but as expected, exhibit significantly weaker performance towards character-level typos. MinHash and SimHash struggle more with word-level augmentations than deep-learning based embeddings and collapse when character-level typos are introduced. We attribute \rs's resilience to adversarial manipulations to the RETVec character encoder as well as using deep metric learning to train robust embeddings.

Figure~\ref{fig:eval:text_aug} reports the Recall@1 performance of the algorithms as the amount of augmentation increases. All algorithms perform perfectly when no augmentation is applied (exact matching), but as the percentage of augmentation increases, n-gram based approaches exhibit a steep drop in performance. Semantic text embeddings are able to sustain a larger degree of augmentation before their retrieval capabilities start to degrade (over 20\%). \rnd is the most robust algorithm, with a noticeable drop in performance only after around 40\% augmentation. This makes \rs the most effective approach at clustering and deduplicating text under adversarial settings.
%  It is this resilience under very adversarial condition that gives \rnd its edge and make it the most effective algorithm to perform tasks under adversarial settings such as clustering spam messages as discussed in detail in Section~\ref{?}.

\vspace{-1mm}

\begin{table}[h]
\begin{minipage}[b]{0.41\textwidth}
\centering
\includegraphics[width=\textwidth]{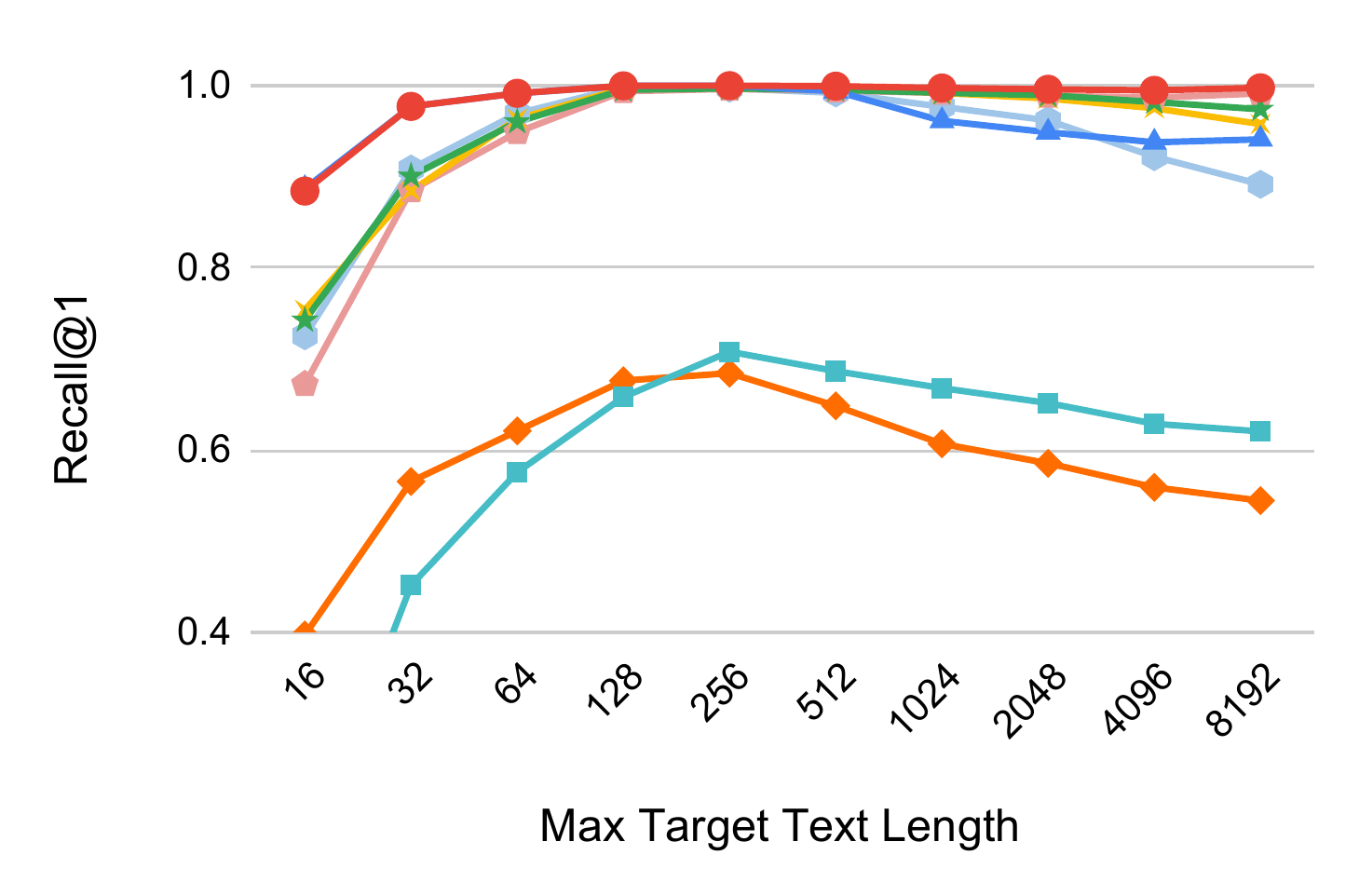}
\vspace{-2mm}
\captionof{figure}{Recall@1 performances on the \wa benchmark (English only) for varying max target lengths.}
\label{fig:eval:target_length}
\end{minipage}\hfill
\begin{minipage}[b]{0.56\linewidth}
\centering
\includegraphics[width=\textwidth]{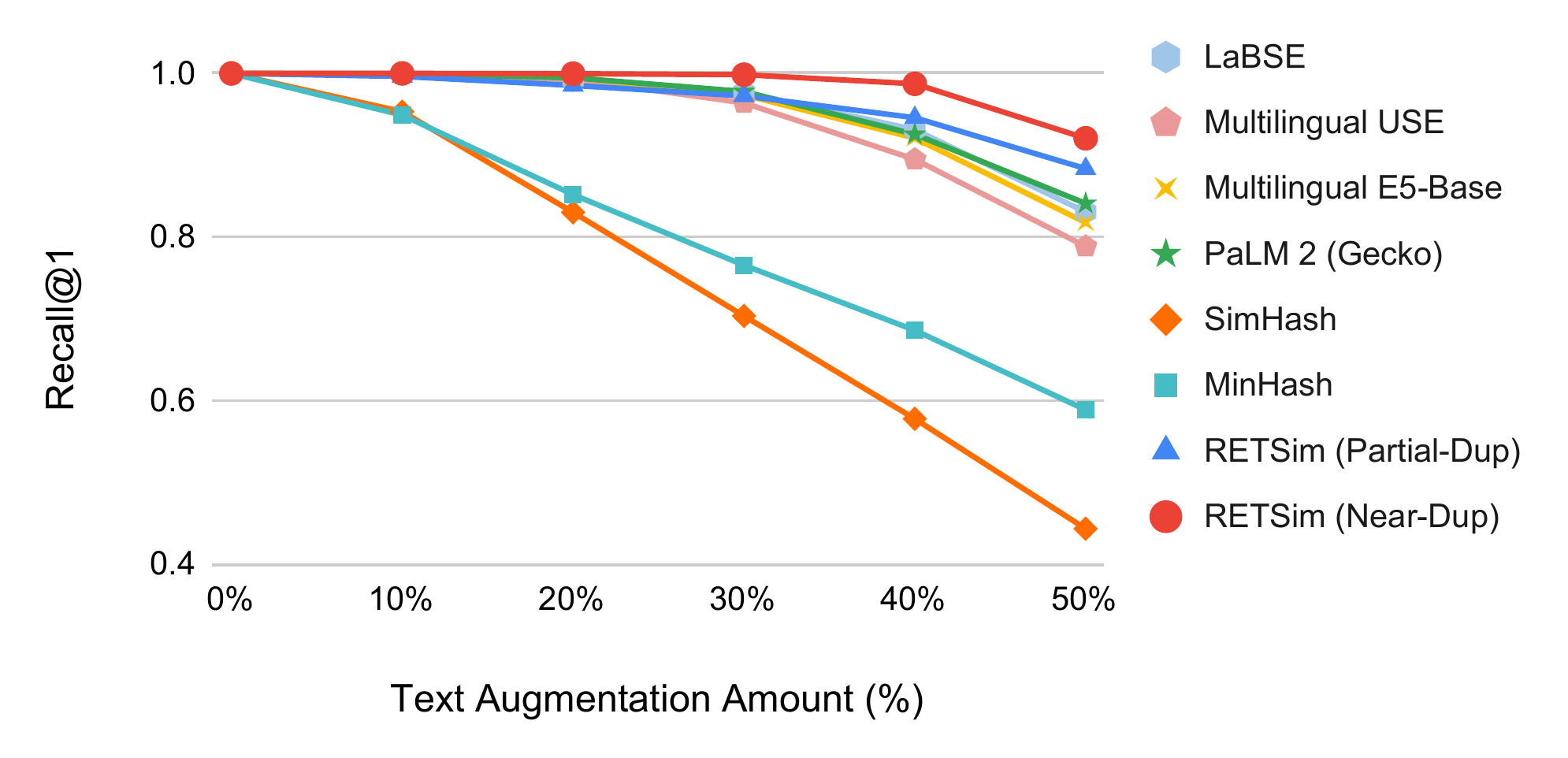}
\captionof{figure}{Recall@1 performances on the \wa benchmark (English only) as the amount of augmentation applied to the query text increases.}
\label{fig:eval:text_aug}
\end{minipage}
\vspace{-2mm}
\end{table}

% \vspace{-2mm}
\paragraph{Text Length Impact on Performance} Figure~\ref{fig:eval:target_length} reports the Recall@1 performance of \rs and baseline algorithms as the length of the query and target text varies. We see that \rnd and \rpd outperforms all other methods on short texts with fewer than 128 characters. As the text length increases beyond 512 characters, \rnd remains close to perfect while \rpd's performance degrades since it splits the text into multiple embeddings and finds the nearest matching \textit{chunk} of text. MinHash and SimHash also perform poorly on short text lengths and start to degrade on longer texts. For neural-based embeddings, we observe a slight drop in performance on longer texts for all models except \rnd and Multilingual USE, the only two embeddings that can handle arbitrary length inputs.

\label{sec:emp:ret}
%\subsubsection{No augmentation with ratio variation}
%We select queries of varying size.
%We perform text retrieval
%\fixme{Text retrieval evaluation no augmentation with ratio variation}
%We evaluate the performance of selected syntactic and semantic algorithms on text retrieval task. We see that for varying query and target sizes the performance of the algorithms vary. Moreover, our observation also shows that the effect of the changing query, text size is not the same for short to long or long to short. 
%\fixme{Add section with MTEB retrieval}
% \subsection{Near-Duplicate Detection Evaluation}
% \label{sec:neardup}

\subsection{Real-World Near-Duplicate Detection Evaluation}
\label{sec:dedup}

\paragraph{Setup}

We benchmark \rs's ability to identify near-duplicate content on real-world datasets from the literature. The NEWS-COPY Deduplication dataset~\citep{silcock_noise-robust_2022} contains 27,210 historical news articles with 122,876 positive duplicate pairs. The dataset consists of noisy near-duplicates due to factors like OCR errors, plagiarism, and news aggregation. We also evaluate the algorithms on the CORE Near-Duplicates dataset~\citep{gyawali_deduplication_2020}, which consists of 100k scholarly articles with 25k exact duplicates, 25k near-duplicates, and 50k non-duplicates. Near-duplicates in this dataset arise from article revisions, versioning and metadata differences, and human typos. A key difference between these two benchmarks and the \wa benchmark is that these two benchmarks are focused on detecting and clustering near-duplicate text, rather than robust text retrieval based on syntactic similarity. For both benchmarks, we follow the experimental setup provided in the papers and report Adjusted Rand Index (ARI) for the NEWS-COPY dataset and report precision/recall/F1 scores on the CORE Near-Duplicates dataset.
 
%  For each of those we follow the papers methodology and report RetSim numbers. Additionally for the XX datasets we also benchmark other neural embeddings to confirm that MinHash outperforms them - for space sake and reduce carbon footprint didn't repeat those uncessary benchmark on the second one.

% \fixme{@marina: please explain that this benchmark is heavily different from the wanted one because here we are looking at partial matches}

% The NEWS-COPY Deduplication dataset is a dataset containing xxx articles~\citep{silcock_noise-robust_2022} from historical .
% Historical OCR causes many typos, etc, to appear in the dataset 
% basically summarize the thing

\begin{table}[h]
\vspace{-2mm}
\begin{minipage}[b]{0.45\linewidth}
\centering
\begin{tabular}{@{}l|rl@{}}
\toprule
% \multicolumn{1}{c}{\textbf{Model/Algorithm}} & \multicolumn{1}{l}{\textbf{ARI}} &  \\ \midrule
\multicolumn{1}{c|}{\textbf{Model/Algorithm}} & \multicolumn{1}{l}{\textbf{ARI}} &  \\ \midrule
Multilingual USE & 0.730 &  \\
Multilingual E5-Base & 0.742 &  \\
S-BERT* & 0.700 &  \\ 
 \midrule
SimHash & 0.695 &  \\
MinHash* & 0.737 &  \\
MinHash (Ours) & \underline{0.783} &  \\ 
 \midrule
\rpd & \textbf{0.831} &  \\
\rnd & 0.704 &  \\ \bottomrule
\end{tabular}
    \caption{Performance comparison on the NEWS-COPY dataset. Adjusted Rand Index (ARI) values are reported. * denotes results from \cite{silcock_noise-robust_2022}.}
    \label{tab:news_copy_ari}
\end{minipage}\hfill
\begin{minipage}[b]{0.5\textwidth}
\centering
\includegraphics[width=\textwidth]{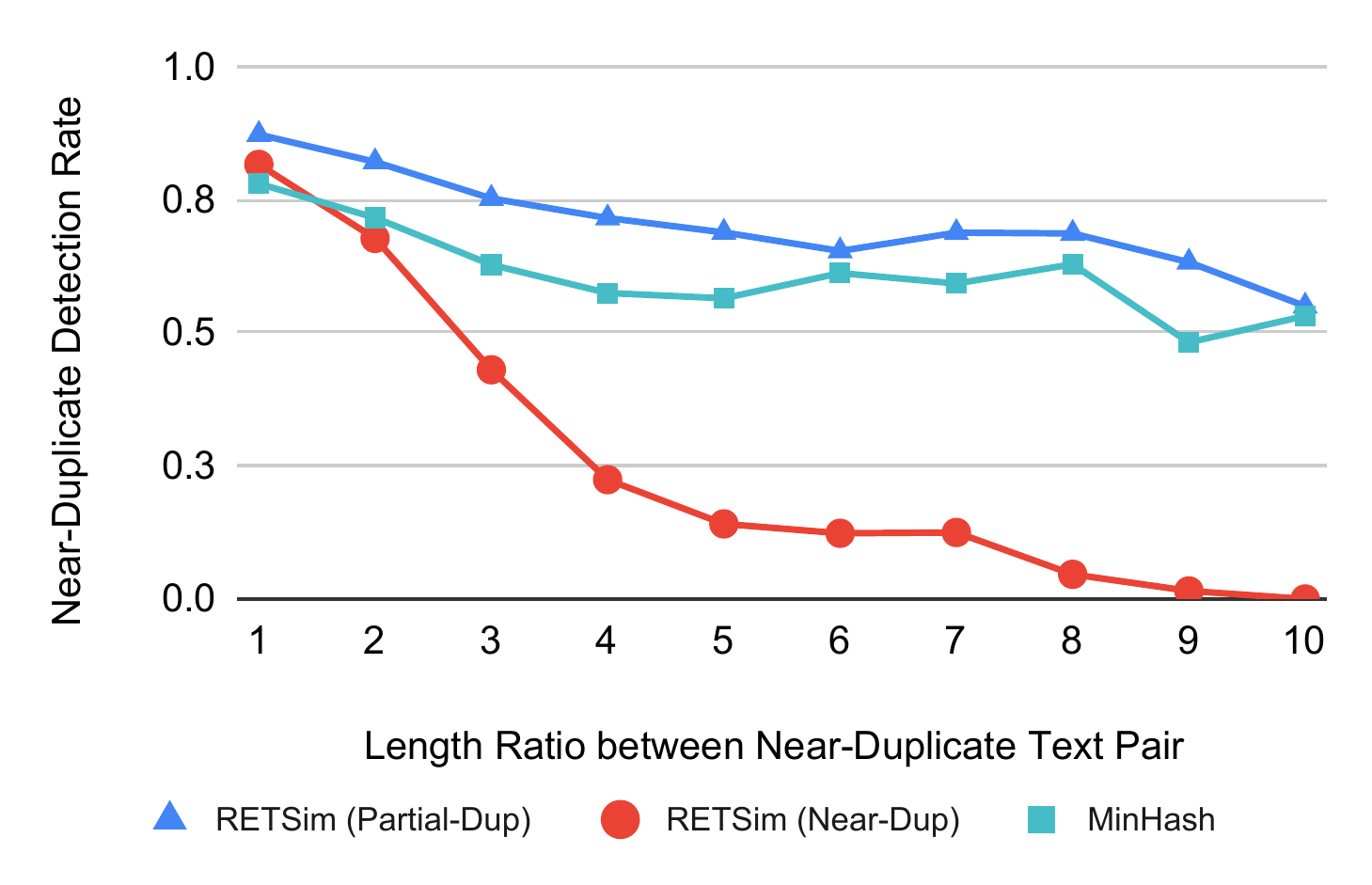}
\captionof{figure}{Near-duplicate detection rate of \rs vs MinHash for different length ratios of positive pairs. X-axis is the length of longer divided by shorter text, rounded to the nearest integer.}
\label{fig:length_ratio_news_copy}
\end{minipage}
\vspace{-2mm}
\end{table}

\paragraph{Results} On the NEWS-COPY dataset, \rpd outperforms all other approaches by a significant margin (4.8\% ARI compared to our best MinHash result), as reported in Table~\ref{tab:news_copy_ari}. In the dataset, there are many near-duplicate pairs where one text is significantly longer than the other, so it is expected that \rpd, which can find matching text chunks in documents, is more suited for the task and outperforms \rnd. Bucketing the near-duplicate detection rate of each algorithm by the length ratio between positive pairs (Figure~\ref{fig:length_ratio_news_copy}), we observe that \rpd outperforms MinHash regardless of the length ratio, but MinHash surpasses \rnd performance when one text is above roughly 1.5x the length of the other text in a near-duplicate pair. Additionally, we noticed that the labels in the dataset were occasionally noisy, as a substantial portion of the \rs false positives appear to be near-duplicates upon inspection (Appendix~\ref{app:select_examples}).

%As expected, as the difference in length increases, \rnd's detection rate drops because irrelevant information in the full text overshadows the near-duplicate parts of the text.

On the CORE Near-Duplicates dataset (Table~\ref{tab:scholar}), where documents (article title + abstract) are roughly the same size, \rpd and \rnd performance is roughly equivalent. Both methods outperform the baselines in terms of macro F1 score and accuracy. We use MinHash + LSH with 256 hash functions for computational efficiency, as recommended by the datasketch library\footnote{datasketch: Big Data Looks Small. https://github.com/ekzhu/datasketch.} for better accuracy than the default setting. Deduplication thresholds and detailed hyperparameter settings for the algorithms on both near-duplication datasets can be found in Appendix~\ref{app:eval_hyp}.

\begin{table}[ht]
\centering
\scalebox{0.75}{\begin{tabular}{@{}l|cccccc@{}}
\toprule
 \multirow{2}{*}{\textbf{Model / Algorithm}} & \multicolumn{1}{l}{\textbf{\begin{tabular}[c]{@{}c@{}}Precision \\ Duplicates\end{tabular}}} & \multicolumn{1}{l}{\textbf{\begin{tabular}[c]{@{}c@{}}Recall \\ Duplicates\end{tabular}}} & \multicolumn{1}{l}{\textbf{\begin{tabular}[c]{@{}c@{}}Precision \\ Non-Duplicates\end{tabular}}} & \multicolumn{1}{l}{\textbf{\begin{tabular}[c]{@{}c@{}}Recall \\ Non-Duplicates\end{tabular}}} & \multicolumn{1}{l}{\textbf{Macro F1}} & \multicolumn{1}{l}{\textbf{Accuracy}} \\ \midrule
Exact Title Matching* & 0.830 & 0.500 & 0.709 & \textbf{0.992} & 0.757 & 0.746 \\ \midrule
% Hybrid Method* & 0.908 & 0.83 & 0.899 & \underline{0.979} & 0.904 & 0.903 \\
LaBSE&	\underline{0.937}&	0.923&	0.930&	0.943&	0.933&	0.919 \\ 
Multilingual USE&	0.917	&0.907&	0.918	&0.927	&0.917&	0.909 \\

Multilingual E5-Base &	0.931&	0.908&	0.919&	0.939	&0.924&	0.920\\ \midrule
MinHash + LSH & 0.929 & 0.902 & 0.915 & 0.938 & 0.921 & 0.918 \\   \midrule
\rpd & \textbf{0.945} & \textbf{0.941} & \textbf{0.945} & \underline{0.949} & \textbf{0.945} & \textbf{0.928} \\
\rnd & 0.928 & \underline{0.937} & \underline{0.942} & 0.934 & \underline{0.935} & \underline{0.926} \\ \bottomrule
\end{tabular}}
\caption{Evaluation results on the CORE Near-Duplicates dataset. Precision/recall/macro F1 and accuracy numbers are reported. * denotes results from~\cite{gyawali_deduplication_2020}.}
\label{tab:scholar}
\end{table}

%% file: applications.tex
\vspace{-2mm}
\section{Applications}
\label{sec:application}

% \fixme{hard to write this section without knowing wikipedia results}
% We benchmark performance and speed on large training datasets, which has been shown to improve training performance of models. 

\subsection{Training Dataset Deduplication}
\label{sec:dataset_deup}

\begin{table}[h]
\centering
\scalebox{0.8}{\begin{tabular}{@{}lrr@{}}
\toprule
\textbf{Model/Algorithm} & \multicolumn{1}{l}{\textbf{\begin{tabular}[c]{@{}l@{}}\% train examples\\  with dup in train\end{tabular}}} & \multicolumn{1}{l}{\textbf{\begin{tabular}[c]{@{}l@{}}\% valid examples\\ with dup in train\end{tabular}}} \\ \midrule
MinHash + LSH & 0.47\% & 0.46\% \\
Exact Substring* & 2.76\% & 0.52\% \\
\rnd & 3.17\% & 0.59\% \\
\rpd & 12.77\% & 2.66\% \\ \bottomrule
\end{tabular}}
\caption{Deduplication rate on Wiki-40B (English). * denotes results from~\cite{lee_deduplicating_2022}.}
\label{tab:wiki_dedup}
\end{table}

\begin{table}[h]
\centering
\scalebox{0.75}{
\begin{tabular}{@{}l|lccc@{}}
\toprule
\multirow{2}{*}{\textbf{Model/Algorithm}} & \multicolumn{1}{c}{\multirow{2}{*}{\textbf{Accelerator}}} & \multicolumn{1}{c}{\multirow{2}{*}{\textbf{Batch Size}}} & \multicolumn{2}{c}{\textbf{Embedding / Hashing}} \\
 & \multicolumn{1}{c}{} & \multicolumn{1}{c}{} & \textbf{time (sec)} & \textbf{examples/sec} \\ \midrule
MinHash + LSH & CPU AMD 7950 32 cores & \multicolumn{1}{c}{-} & 234 & 12544 \\
RETSim & Onnx CPU AMD 7950 32 cores & 256 & 10839 & 270 \\
RETSim & TensorFlow GPU RTX 4090 & 4096 & 720 & 4062 \\
RETSim & TensorFlow GPU NVIDIA H100 & 16384 & 363 & 8069 \\ \bottomrule
\end{tabular}}
\caption{Embedding/hashing speed of RETSim vs MinHash + LSH on the Wiki-40B dataset.}
\vspace{-3mm}
\label{tab:dedup_speed}
\end{table}

% \subsubsection{Setup}
\vspace{-3mm}

\paragraph{Setup} We evaluate \rs's ability to deduplicate text training datasets by deduplicating the English split of Wiki-40B~\citep{guo_wiki-40b_2020}. We conservatively set the cosine similarity deduplication threshold to 0.1 for \rnd and 0.15 for \rpd to limit the amount of false positives, based on the optimal thresholds found in the evaluation (Appendix~\ref{app:eval_hyp}). We use USearch's default vector index for approximate nearest neighbor search~\citep{Vardanian_USearch_by_Unum_2023}. We compare against MinHash + LSH, where we set the number of hash functions to be 256 following~\cite{kocetkov_stack_2022} and use a Jaccard similarity threshold of 0.8 for deduplication~\citep{lee_deduplicating_2022}.

% showed that a threshold between 0.8 and 0.9 ensures. 256 hash functions for MinHash + LSH is also commonly used in the literature for large-scale dataset deduplication

% \subsubsection{Deduplication Performance}

% \vspace{-2mm}
\paragraph{Results} Overall, as reported in Table~\ref{tab:wiki_dedup}, \rnd finds slightly more duplicates in the Wiki-40B training and validation splits. This is in-line with our deduplication results (Section~\ref{sec:dedup}) where \rnd outperforms other algorithms. On the other hand, \rpd finds significantly more matches than the exact substring matching algorithm used in the previous study~\citep{lee_deduplicating_2022}, showcasing the usefulness of performing both near-duplicate and partial-duplicate matching at once. This larger-than-expected number of partial matches indicate that machine learning practitioners should take extra care to deduplicate Wikipedia at the chunk level to avoid feeding duplicate text to their models.

In terms of embedding speed (Table~\ref{tab:dedup_speed}), \rs is significantly slower than MinHash + LSH on CPU (46x slower), competitive when using a desktop GPU such as the RTX 4090 (3x slower) and almost on-par when using a high-end GPU like the NVIDIA H100 (1.5x slower). Our current code is written in Python and not fully optimized, so we expect this performance gap to significantly shrink as we optimize our implementation. Although \rs is slower than MinHash, \rs is significantly smaller and faster than other text embedding models, and closes the performance gap between neural and non-neural based methods for near-duplicate text detection and dataset deduplication. Both \rnd and \rpd are returned at the same time so they have the same embedding speed. Indexing and retrieval times will depend on the vector index and search algorithm used. For longer documents, \rpd will produce more embeddings than \rnd, so \rpd offers a tradeoff between finer-grained matching versus indexing/retrieval speed, which will depend on the specific vector search algorithm and dataset used.

\subsection{In the Wild: Spam Email Clustering}
\label{sec:spam_email}
In this section, we showcase \rs's real-world performance on clustering near-duplicate text which has been heavily manipulated by adversarial attacks by performing an evaluation on spam campaigns. Spam constitutes a strong proving ground for near-duplicate clustering algorithms as spammers employ adversarial augmentation techniques in an attempt to evade detection. Such augmentations typically include appending or prepending unrelated text, interleaving random words and different languages, intentionally introducing typos, abusing extended character sets such as emojis and homoglyphs, and more. These techniques are collectively referred to as hash-busting.

\paragraph{Setup}  The dataset consists of 5,252 spam emails from 196 spam campaigns, donated by Gmail users who flagged them when they reached their inboxes. Each example contains the email subject concatenated with the message content. The emails were misclassified by a spam classifier due to their effective adversarial text manipulation techniques, which makes them a challenging test set for clustering evaluations. Some examples of hash-busting attacks and adversarial manipulations we observe include the use of homoglpyphs, uncommon Unicode character sets, invisible characters, and padding with random words from different languages. To get the ground truth campaign clusters, emails were manually reviewed and assigned to a specific spam campaign based on similarity by human reviewers. We use agglomerative clustering to cluster spam emails, and report homogeneity, completeness, V-Measure, and Adjusted Rand Index (ARI) metrics.
 
\begin{table}[ht]
\centering
\scalebox{0.8}{
\begin{tabular}{@{}l|rrrr@{}}
\toprule
\textbf{Model / Algorithm} & \multicolumn{1}{l}{\textbf{Homogeneity}} & \multicolumn{1}{l}{\textbf{Completeness}} & \multicolumn{1}{l}{\textbf{V-Measure}} & \multicolumn{1}{l}{\textbf{ARI}} \\ \midrule
USE & 0.856 & 0.955 & 0.903 & 0.6 \\
SimHash + LSH & 0.867 & 0.876 & 0.871 & 0.571 \\
\rnd & \textbf{0.937} & \textbf{0.963} & \textbf{0.949} & \textbf{0.747} \\ \bottomrule
\end{tabular}}
\caption{Performance on clustering adversarial spam campaigns in practice.}
\vspace{-2mm}

\label{tab:clustering:spam}
\end{table}

\paragraph{Results} Overall, we observed that \rs is significantly better at clustering near-duplicates with adversarial manipulations, outperforming both SimHash and USE across all metrics considered (Table~\ref{tab:clustering:spam}). In particular, we observed that \rs outperforms USE by 4.6\% on the V-Measure score which is our main metric. The results reported in this section are in-line with what we observe since we deployed \rs as our main near-duplicate detection algorithm in December 2022.

%% file: ablation.tex
\section{Ablation Studies}
\label{sec:abl}
\paragraph{Setup} In this section, we summarize the key ablation studies we performed when designing \rs. All the models used in this section are trained using the setup detailed in Appendix~\ref{sec:app:tra}, except we only train them for 100k steps to reduce computational costs. We evaluate \rnd's performance for each model on a subset of the \wa benchmark, where we randomly select 1000 examples from each of the 41 language splits and use Recall@1 as reported metric.

\begin{table}[ht]
\scalebox{0.8}{\begin{tabular}{@{}rrl@{}}
\toprule
\multicolumn{1}{c}{\textbf{Block Type}} & \multicolumn{1}{c}{\textbf{Recall@1}} &  \\ \midrule
RETVec MLP & 0.975 &  \\
ConvNeXt & 0.978 &  \\
BERT & 0.973 &  \\
T5 & 0.980 &  \\
\textit{\textbf{*GAU}} & \textbf{0.986} &  \\ \bottomrule
\end{tabular}}
\hfill
\scalebox{0.8}{\begin{tabular}{@{}rrl@{}}
\toprule
\multicolumn{1}{l}{\textbf{Chunk Size}} & \multicolumn{1}{l}{\textbf{Recall@1}} &  \\ \midrule
128 & 0.979 &  \\
256 & 0.984 &  \\
\textit{\textbf{*512}} & \textbf{0.986} &  \\
1024 & 0.983 &  \\
2048 & 0.978 &  \\ \bottomrule
\end{tabular}}
\hfill
\scalebox{0.8}{\begin{tabular}{@{}rrl@{}}
\toprule
\multicolumn{1}{c}{\textbf{Embed. Dim}} & \multicolumn{1}{c}{\textbf{Recall@1}} &  \\ \midrule
64 & 0.969 &  \\
128 & 0.980 &  \\
\textit{\textbf{*256}} & \textbf{0.986} &  \\
512 & 0.986 &  \\
768 & 0.986 &  \\ \bottomrule
\end{tabular}
}
\hfill
\caption{\rs ablation study results on architecture block type (left), text chunk size (middle), and embedding dimension (right). \textbf{*\textit{Bold}} denotes the value selected for the final \rs model.}
\vspace{-1mm}
 \label{tab:abl}
\end{table}

\vspace{-1mm}
\paragraph{Results}  
Table~\ref{tab:abl} contains RETSim ablation study results on max text chunk size, architecture block type, and embedding size. The most important architectural decision was to decide the optimal text chunk size and finding the right balance between having the smallest size possible to maximize \rpd efficiency while ensuring \rnd full-text embeddings can work effectively on full documents. We find that chunks of 512 characters offer the best performance.

We also tested various model architectures and transformer blocks to find the best balance between efficiency and performance. We find that the more modern GAU block~\citep{hua_transformer_2022} outperforms the vanilla BERT transformer block~\citep{devlin_bert_2019} and the T5 block~\citep{xue_mt5_2020}. We also tried modern CNN architectures such as ConvNeXt~\citep{liu_convnet_2022} and the MLP architecture proposed in RETVec~\citep{bursztein_retvec_2023}, but both were worse than GAU block performance. Last but not least, we found that increasing the embedding size past 256 dimensions does not yield any meaningful improvements for \rnd. Accordingly, we opted to use a 256-dimension embedding for space-efficiency and to maximize indexing and query speed. Additional ablation studies for other hyperparameters can be found in Appendix~\ref{app:ablation}.
 
% \begin{table}[ht]
% \begin{tabular}{|l|c|c|c|c|c|}
%  \hline
%   & \multicolumn{5}{c|}{Word + Character Typo Percentage}  \\
% Pooling Type &0\% & 10\% & 20\% & 30\% & Avg\\
% \hline
% \hline

% Avg         & 100.00\% & 92.29\%          & 88.29\%          & 86.86\%          & 91.86\%        \\
% GeM (p=2)   & 100.00\% & 92.57\%          & 89.00\%          & 87.14\%          & 92.18\%           \\
% GeM (p=3) * & 100.00\% & \underline{93.57}\%    & \textbf{89.71}\% & 87.43\%          & \textbf{92.68}\% \\
% GeM (p=5)   & 100.00\% & 93.29\%          & \underline{89.57}\%    & \textbf{87.71}\% & \underline{92.64}\%       \\
% Max         & 100.00\% & \textbf{93.86}\% & 88.86\%          & \underline{87.57}\%    & 92.57\%           \\

% \hline
% \end{tabular}
% \caption{RetSim pooling type ablation study on the multilingual Wiki-40B retrieval benchmark at various augmentation levels. * denotes the value selected for the final model.}
% \label{tab:abl:poo}
% \end{table}

%% file: conclusion.tex
\section{Future Work}
% \rs achieves new state-of-the-art performance on near-duplicate detection even under adversarial conditions at a fraction of the size of other neural embeddings. This performance/efficiency ratio is achieved at the expense of not being as competitive as larger models on semantic tasks.

%In that sense, \rs is very similar to the YOLO models that specialize in real-time object detection~\citep{redmonYouOnlyLook2016a, redmonYOLO9000BetterFaster2017, redmonYOLOv3IncrementalImprovement2018}

%There are potentially many other areas where making the concious tradeoff between performance and generality will yield practical benefits, with image deduplication being a prime candidate that we plan on exploring next.
%\rs uses a novel training regime which combines metric learning with data augmentation
\vspace{-1mm}
\rs's novel training regime, which combines metric learning and data augmentation, has many other potential applications that we plan to explore in future work. For example, it could be adapted or extended to train robust semantic embeddings or image similarity embeddings. Additionally, we expect that as general models become bigger and more expensive to run in the future, smaller, specialized models such as \rs will emerge as an efficient alternative for a wide range of tasks.

\section{Conclusion}
\label{sec:con}
\vspace{-1mm}

In this paper, we introduced \rs, a novel, multilingual text embedding which achieves state-of-the-art performance on near-duplicate text detection, dataset deduplication, and syntactic text similarity benchmarks. \rs is significantly faster than previous neural-based text embeddings and more robust than n-gram based algorithms, which makes it suitable for large-scale text retrieval and dataset deduplication, especially in adversarial settings such as spam detection. Furthermore, we introduced the \wa benchmark, the first multilingual dataset designed to measure the adversarial robustness of near-duplicate text detection algorithms. We open-source both \rs and the \wa benchmark under the MIT License.

%% file: appendix.tex
\section{Appendix}
\label{sec:app}

\subsection{\rs Details}
\label{sec:app:mdl}

\subsubsection{\rs Model Hyperparameters}
\label{sec:app:mdl:hyp}

The full list of RETSim model hyperparameters can be found in Table~\ref{tab:app:mdl}.

\begin{table}[H]
    \centering
\begin{tabular}{@{}l|rr@{}}
\toprule
\textbf{Hyperparameter} & \textbf{Value} &  \\ \midrule
Max input length (per chunk) & \multicolumn{1}{r}{512} &  \\
Block type & GAU &  \\
\# blocks & \multicolumn{1}{r}{2} &  \\
Hidden dim & \multicolumn{1}{r}{256} &  \\
Expansion rate & \multicolumn{1}{r}{1} &  \\
Activation function & Swish &  \\
Attention activation function & relu$^2$ &  \\
Absolute positional encoding & ScaledSin &  \\
Relative positional encoding & RoPE &  \\
Norm type & ScaleNorm &  \\
Pooling type & GeM $(p=3)$ &  \\
Dropout rate & \multicolumn{1}{r}{0} &  \\
Embedding dim & \multicolumn{1}{r}{256} &  \\
\# Parameters & 536k &  \\ \bottomrule
\end{tabular}
\caption{Detailed \rs model hyperparameters.}
\label{tab:app:mdl}
\end{table}

\subsubsection{\rs Training Hyperparameters}
\label{sec:app:tra}
Table~\ref{tab:app:tra} details the hyperparameters settings for training configuration, loss, and optimizer used to train the \rs model.

\begin{table}[H]
    \centering
\begin{tabular}{@{}l|rr@{}}
\toprule
\textbf{Hyperparameter} & \multicolumn{1}{l}{\textbf{Value}} &  \\ \midrule
Batch size & 1024 &  \\
Train steps & \multicolumn{1}{l}{1 million} &  \\
LAMB $\epsilon$ & 1e-6 &  \\
LAMB $\beta_1$ & 0.9 &  \\
LAMB $\beta_2$ & 0.999 &  \\
Max learning rate & 0.001 &  \\
End learning rate & 0 &  \\
Learning rate decay & Cosine &  \\
Weight decay & 0 &  \\ \bottomrule
\end{tabular}
\caption{\rs detailed training hyperparameters.}
\label{tab:app:tra}
\end{table}

% \subsection{Additional Benchmarking on XLSUM, The stack-smol programming languages database}
% \label{sec:app:new}

\subsection{Training Dataset Details}
\label{sec:app:aug}

Below, we provide the full list of augmentations used to generate augmented text for the \rs training dataset, as described in Section~\ref{sec:training_dataset}.

\subsubsection*{Sentence-level augmentations}

\begin{itemize}
    \item Deletion:
        \begin{itemize}
            \item Random sentence deletion
            \item Random sentence truncation
        \end{itemize}
    \item Insertion:
        \begin{itemize}
            \item Random prefix sentence
            \item Random suffix sentence
            \item Random sentence insertion
            \item Repeat sentence
        \end{itemize}
    \item Substitution:
        \begin{itemize}
            \item Lowercase/uppercase sentence
            \item Random sentence substitution
        \end{itemize}
    \item Transposition:
        \begin{itemize}
            \item Neighboring Swap
        \end{itemize}
\end{itemize}

\subsubsection*{Word-level augmentations}
\begin{itemize}
    \item Deletion:
        \begin{itemize}
            \item Random word deletion
        \end{itemize}
    \item Insertion:
        \begin{itemize}
            \item Random word insertion
            \item Random word insertion per language
        \end{itemize}
    \item Substitution:
        \begin{itemize}
            \item 3-gram frequency based word substitution
            \item Random word substitution
            \item Random word substitution per language
            \item Repeat word
        \end{itemize}
    \item Transposition:
        \begin{itemize}
            \item Neighboring Swap
        \end{itemize}
\end{itemize}

\subsubsection*{Character-level augmentations}
\begin{itemize}
    \item Deletion:
        \begin{itemize}
        \item Random character deletion
        \end{itemize}
    \item Substitution:
        \begin{itemize}
        \item Case substitution
        \item n-gram based substitution for $n=3, 4, 5$
        \item QWERTY keyboard typo substitution
        \item Homoglyphs substitution
        \item Random ASCII substitution
        \item Random character from language alphabet substitution
        \item Random punctuation substitution
        \item Random Unicode character substitution
        \end{itemize}
    \item Insertion:
    \begin{itemize}
    \item Character repetition
    \item n-grams based insertion for $n=3, 4, 5$
    \item Random character from language alphabet insertion
    \item Random punctuation insertion
    \item Random Unicode character insertion
    \end{itemize}
    \item Transposition:
        \begin{itemize}
        \item Neighboring swap
        \end{itemize}
\end{itemize}

\subsection{Detailed Evaluation Hyperparameters}
\label{app:eval_hyp}

Figures~\ref{fig:app:news_hyp} and~\ref{fig:app:core_hyp} contain information on deduplication thresholds values and hyperparameter settings for each algorithm benchmarked on the NEWS-COPY and CORE deduplication datasets.

\begin{figure}[h]
\centering
\scalebox{0.9}{\begin{tabular}{@{}llrl@{}}
\toprule
\textbf{Model / Algorithm} & \textbf{Threshold Type} & \multicolumn{1}{l}{\textbf{Threshold Value}} & \textbf{Hyperparameters} \\ \midrule
Multilingual USE & Cosine Similarity & 0.96 & - \\
Multilingual E5-Base & Cosine Similarity & 0.88 & - \\
SimHash & Hamming Distance & 10 & 64 bits, 5-grams (character-level) \\
MinHash (Ours) & Jaccard Similarity & 0.6 & 10 hash functions, 2-grams (word-level) \\
\rnd & Cosine Similarity & 0.89 & - \\
\rpd & Cosine Similarity & 0.84 & - \\ \bottomrule
\end{tabular}}
    \caption{Hyperparameter settings for NEWS-COPY dataset evaluation in Section~\ref{sec:dedup}.}%
    \label{fig:app:news_hyp}%
\end{figure}

\begin{figure}[h]
\centering
\scalebox{0.9}{\begin{tabular}{@{}llrl@{}}
\toprule
\textbf{Model / Algorithm} & \textbf{Threshold Type} & \multicolumn{1}{l}{\textbf{Threshold Value}} & \textbf{Hyperparameters} \\ \midrule
LaBSE  & Cosine Similarity & 0.88 & - \\
Multilingual USE & Cosine Similarity & 0.97 & - \\
Multilingual E5-Base & Cosine Similarity & 0.87 & - \\
SimHash + LSH & Hamming Distance & 6 & 64 bits, 3-grams (character-level) \\
MinHash + LSH & Jaccard Similarity & 0.5 & 256 hash functions, 3-grams (word-level) \\
\rnd & Cosine Similarity & 0.86 & - \\
\rpd & Cosine Similarity & 0.82 & - \\ \bottomrule
\end{tabular}}
    \caption{Hyperparameter settings for CORE Near-Duplicates dataset evaluation in Section~\ref{sec:dedup}.}%
    \label{fig:app:core_hyp}%
\end{figure}

\subsubsection{Deduplication Threshold Impact}
\label{app:threshold}

\begin{figure}[h]
\centering
\scalebox{0.9}{
    \centering
   \centering{\includegraphics[width=0.47\textwidth]{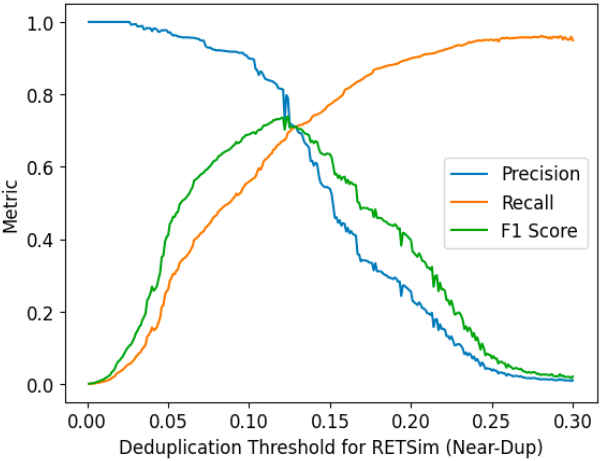} }%
    \qquad
    \centering{\includegraphics[width=0.47\textwidth]{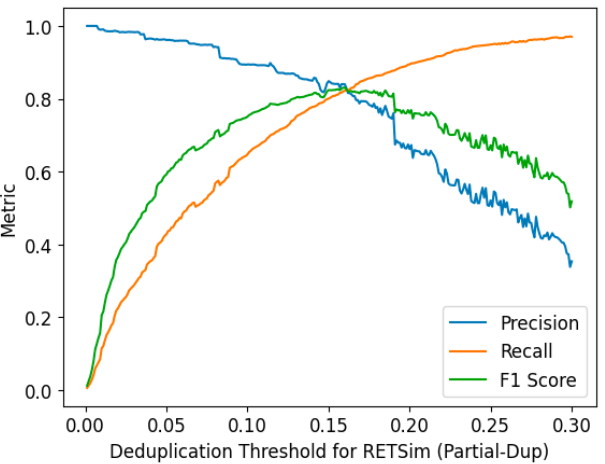} }}%
    \caption{Precision/Recall/F1 scores for different cosine distance deduplication thresholds for \rnd (left) and \rpd (right) on the NEWS-COPY dataset.}%
    \label{fig:app:dedup_threshold}%
\end{figure}

\subsection{Detailed \wa Benchmark Results}
\label{app:wiki_full}

Tables~\ref{fig:app:wiki_full_1} and~\ref{fig:app:wiki_full_2} show detailed performance results for \rs and all baseline algorithms for every language split in the \wa benchmark.

\begin{landscape} 
\begin{table}[h]
\centering
\scalebox{0.75}{
\centering
\begin{tabular}{@{}lrrrrrrrrrrrrrrrrrrrrr@{}}
\toprule
\textbf{Model / Algorithm} & \multicolumn{1}{l}{\textbf{ar}} & \multicolumn{1}{l}{\textbf{bg}} & \multicolumn{1}{l}{\textbf{ca}} & \multicolumn{1}{l}{\textbf{cs}} & \multicolumn{1}{l}{\textbf{da}} & \multicolumn{1}{l}{\textbf{de}} & \multicolumn{1}{l}{\textbf{el}} & \multicolumn{1}{l}{\textbf{en}} & \multicolumn{1}{l}{\textbf{es}} & \multicolumn{1}{l}{\textbf{et}} & \multicolumn{1}{l}{\textbf{fa}} & \multicolumn{1}{l}{\textbf{fi}} & \multicolumn{1}{l}{\textbf{fr}} & \multicolumn{1}{l}{\textbf{he}} & \multicolumn{1}{l}{\textbf{hi}} & \multicolumn{1}{l}{\textbf{hr}} & \multicolumn{1}{l}{\textbf{hu}} & \multicolumn{1}{l}{\textbf{id}} & \multicolumn{1}{l}{\textbf{it}} & \multicolumn{1}{l}{\textbf{ja}} & \multicolumn{1}{l}{\textbf{ko}} \\ \midrule
LaBSE & 0.915 & 0.915 & 0.897 & 0.912 & 0.938 & 0.931 & 0.918 & 0.944 & 0.888 & 0.923 & 0.912 & 0.926 & 0.930 & 0.937 & 0.927 & 0.912 & 0.898 & 0.915 & 0.929 & 0.931 & 0.949 \\
Multilingual USE & 0.915 & 0.909 & 0.883 & 0.900 & 0.927 & 0.942 & 0.874 & 0.958 & 0.903 & 0.930 & 0.870 & 0.913 & 0.938 & 0.841 & 0.648 & 0.889 & 0.885 & 0.928 & 0.937 & 0.990 & 0.984 \\
Multilingual E5-Base & 0.936 & 0.899 & 0.890 & 0.927 & 0.956 & 0.944 & 0.934 & 0.959 & 0.896 & 0.951 & 0.929 & 0.943 & 0.948 & 0.958 & 0.964 & 0.901 & 0.889 & 0.942 & 0.933 & 0.979 & 0.986 \\
PaLM 2 (Gecko) & 0.497 & 0.851 & 0.902 & 0.911 & 0.943 & 0.932 & 0.589 & 0.961 & 0.911 & 0.926 & 0.356 & 0.914 & 0.934 & 0.589 & 0.435 & 0.903 & 0.876 & 0.924 & 0.937 & 0.578 & 0.701 \\
SimHash & 0.558 & 0.510 & 0.485 & 0.579 & 0.592 & 0.561 & 0.557 & 0.591 & 0.496 & 0.665 & 0.568 & 0.632 & 0.519 & 0.651 & 0.563 & 0.557 & 0.530 & 0.533 & 0.513 & 0.465 & 0.593 \\
MinHash & 0.633 & 0.506 & 0.508 & 0.581 & 0.598 & 0.558 & 0.574 & 0.591 & 0.504 & 0.585 & 0.595 & 0.568 & 0.556 & 0.697 & 0.606 & 0.556 & 0.512 & 0.563 & 0.575 & 0.223 & 0.814 \\
\rpd & 0.928 & 0.942 & 0.924 & 0.950 & 0.953 & 0.949 & 0.935 & 0.954 & 0.938 & 0.953 & 0.946 & 0.956 & 0.947 & 0.945 & 0.941 & 0.934 & 0.950 & 0.950 & 0.944 & 0.963 & 0.971 \\
\rnd & 0.971 & 0.976 & 0.968 & 0.973 & 0.988 & 0.978 & 0.974 & 0.987 & 0.976 & 0.985 & 0.971 & 0.981 & 0.983 & 0.989 & 0.975 & 0.962 & 0.962 & 0.976 & 0.982 & 0.986 & 0.991 \\ \bottomrule
\end{tabular}}
    \caption{Full per-language Recall@1 performance for various embedding models and algorithms on the \wa benchmark (part 1).}%
    \label{fig:app:wiki_full_1}%
\end{table}

\vspace{3cm}
\begin{table}[h]
\centering
\scalebox{0.75}{
\centering
\begin{tabular}{@{}lrrrrrrrrrrrrrrrrrrrr@{}}
\toprule
\textbf{Model / Algorithm} & \multicolumn{1}{l}{\textbf{lt}} & \multicolumn{1}{l}{\textbf{lv}} & \multicolumn{1}{l}{\textbf{ms}} & \multicolumn{1}{l}{\textbf{nl}} & \multicolumn{1}{l}{\textbf{no}} & \multicolumn{1}{l}{\textbf{pl}} & \multicolumn{1}{l}{\textbf{pt}} & \multicolumn{1}{l}{\textbf{ro}} & \multicolumn{1}{l}{\textbf{ru}} & \multicolumn{1}{l}{\textbf{sk}} & \multicolumn{1}{l}{\textbf{sl}} & \multicolumn{1}{l}{\textbf{sr}} & \multicolumn{1}{l}{\textbf{sv}} & \multicolumn{1}{l}{\textbf{th}} & \multicolumn{1}{l}{\textbf{tl}} & \multicolumn{1}{l}{\textbf{tr}} & \multicolumn{1}{l}{\textbf{uk}} & \multicolumn{1}{l}{\textbf{vi}} & \multicolumn{1}{l}{\textbf{zh-cn}} & \multicolumn{1}{l}{\textbf{zh-tw}} \\ \midrule
LaBSE & 0.919 & 0.922 & 0.919 & 0.931 & 0.928 & 0.928 & 0.944 & 0.909 & 0.918 & 0.922 & 0.931 & 0.930 & 0.906 & 0.882 & 0.947 & 0.930 & 0.899 & 0.932 & 0.917 & 0.918 \\
Multilingual USE & 0.902 & 0.919 & 0.932 & 0.936 & 0.921 & 0.931 & 0.952 & 0.869 & 0.910 & 0.901 & 0.908 & 0.906 & 0.899 & 0.888 & 0.949 & 0.940 & 0.893 & 0.910 & 0.986 & 0.985 \\
Multilingual E5-Base & 0.941 & 0.935 & 0.949 & 0.934 & 0.944 & 0.928 & 0.955 & 0.936 & 0.911 & 0.921 & 0.940 & 0.896 & 0.925 & 0.921 & 0.969 & 0.951 & 0.872 & 0.945 & 0.980 & 0.978 \\
PaLM 2 (Gecko) & 0.909 & 0.907 & 0.928 & 0.931 & 0.930 & 0.913 & 0.950 & 0.893 & 0.851 & 0.924 & 0.916 & 0.856 & 0.919 & 0.571 & 0.944 & 0.902 & 0.822 & 0.863 & 0.623 & 0.609 \\
SimHash & 0.609 & 0.624 & 0.533 & 0.527 & 0.577 & 0.586 & 0.548 & 0.514 & 0.554 & 0.575 & 0.580 & 0.553 & 0.552 & 0.669 & 0.507 & 0.606 & 0.517 & 0.609 & 0.276 & 0.315 \\
MinHash & 0.568 & 0.579 & 0.511 & 0.540 & 0.560 & 0.563 & 0.573 & 0.520 & 0.523 & 0.573 & 0.570 & 0.525 & 0.520 & 0.416 & 0.570 & 0.583 & 0.520 & 0.581 & 0.172 & 0.200 \\
\rpd & 0.957 & 0.945 & 0.961 & 0.950 & 0.958 & 0.953 & 0.954 & 0.948 & 0.946 & 0.961 & 0.947 & 0.963 & 0.953 & 0.941 & 0.961 & 0.954 & 0.941 & 0.951 & 0.946 & 0.957 \\
\rnd & 0.980 & 0.983 & 0.980 & 0.979 & 0.981 & 0.979 & 0.985 & 0.977 & 0.970 & 0.971 & 0.977 & 0.979 & 0.969 & 0.946 & 0.989 & 0.978 & 0.957 & 0.985 & 0.971 & 0.968 \\ \bottomrule
\end{tabular}
}
    \caption{Full per-language Recall@1 performance for various embedding models and algorithms on the \wa benchmark (part 2).}%
    \label{fig:app:wiki_full_2}%
\end{table}

\end{landscape}

\subsection{Additional Ablation Studies}
\label{app:ablation}

This section includes ablation studies on additional hyperparameters for the \rs model, including the loss function, pooling type, and model capacity.  

\begin{table}[h]
\centering
\begin{tabular}{@{}cccc@{}}
\toprule
\multicolumn{1}{l}{\textbf{$\alpha$}} & \multicolumn{1}{l}{\textbf{$\beta$}} & \multicolumn{1}{l}{\textbf{$\lambda$}} & \multicolumn{1}{l}{\textbf{Recall@1}} \\ \midrule
2 & 20 & 0.5 & 0.982 \\
2 & 20 & 1 & 0.948 \\
2 & 40 & 0.5 & 0.984 \\
2 & 40 & 1 & 0.919 \\
4 & 20 & 0.5 & 0.982 \\
4 & 20 & 1 & 0.947 \\
\textbf{4} & \textbf{40} & \textbf{0.5} & \textbf{0.986} \\
4 & 40 & 1 & 0.923 \\ \bottomrule
\end{tabular}
    \caption{Ablation study on Multi-Similarity Loss hyperparameters for \rs training. \textbf{Bold} indicates the hyperparameter setting selected for the final model.}%
    \label{fig:app:abl:multisim}%
\end{table}

\begin{table}[h]
\centering
\begin{tabular}{@{}rrr@{}}
\toprule
\multicolumn{1}{l}{\textbf{\# Blocks}} & \multicolumn{1}{l}{\textbf{Hidden Dim}} & \multicolumn{1}{l}{\textbf{Recall@1}} \\ \midrule
2 & 64 & 0.965 \\
2 & 128 & 0.980 \\
\textbf{2} & \textbf{256} & \textbf{0.986} \\
2 & 512 & 0.986 \\
3 & 64 & 0.962 \\
3 & 128 & 0.980 \\
3 & 256 & 0.984 \\
3 & 512 & 0.987 \\
4 & 64 & 0.966 \\
4 & 128 & 0.980 \\
4 & 256 & 0.985 \\
4 & 512 & 0.986 \\ \bottomrule
\end{tabular}
    \caption{Ablation study for \rs model capacity and size (number of GAU blocks and hidden dimension for the blocks). \textbf{Bold} indicates the hyperparameter setting selected for the final model.}%
    \label{fig:app:abl:size}%
\end{table}

\begin{table}[h]
\centering
\begin{tabular}{@{}lrl@{}}
\toprule
\textbf{Pooling Type} & \multicolumn{1}{l}{\textbf{Recall@1}} &  \\ \midrule
Average Pooling & 0.985 &  \\
Max Pooling & 0.983 &  \\
\textbf{Generalized Mean Pooling} & \textbf{0.986} &  \\ \bottomrule
\end{tabular}
    \caption{Ablation study on pooling type for the \rs model. \textbf{Bold} indicates the hyperparameter setting selected for the final model.}%
    \label{fig:app:abl:pool}%
\end{table}

\subsection{Selected Examples from NEWS-COPY Dataset}
\label{app:select_examples}

In this section, we randomly selected a set of false positives and false negatives for \rs on the NEWS-COPY deduplication dataset to provide further insight into the results. 

\begin{table}[h]
\centering
\begin{tabularx}{\textwidth}{|X|X|}\hline
Text 1 & Text 2 \\ \hline

\RaggedRight{chauffeur, a policeman and a passing journalist who tried to intervene. Beaton and the policeman were reported in serious condition. The 23-year-old princess and her husband of five months, Capt. Mark Phillips, were not hurt. But police experts said the holes left by one of the bullets fired into the car indicated it passed between them, missing them by inch- es. A police informant said it was believed 11 shots were fired by the assailant. Experts were studying two revolvers found at the scene. They said fi...}&
\RaggedRight{‘LONDON (AP) — Ian Ball, a 26-year- old unemployed Englishman, was brought into court today and charged with attempted murder during an at- tempt to kidnap Princess Anne from her car in the heart of London Wed- nesday night. Ball, lean-faced and bearded, stood stiffly in the dock at the Bow Street Magistrate’s court, handcuffed to two detectives. He spoke only once during his 60-second appearance, saying iha London accent: “I want to apply for legal aid.” The court ordered him held for another hearing on Ma...}\\\hline

\RaggedRight{By United Press tnfernational Ay SSAST OR BE FRE NG SG The federal government has proposed new methods of eoustructing federal buildings in a move to save ad- ditional energy and suggested ils elfort could be adapted to all new buildings,}&
\RaggedRight{Hy United Press International The federal government has Proposed new methods of constructing federal buildings in a move lo save addilional energy and suggested ils effort could be adapted to all new buildings, Arthur F, Sampson, General Services Administration ad- ministrater, said new features for such construction would include the collection of rain waler for cooling and irriga- tion, solar energy collectors and the covering of exterior walls with earth. “Whal we are saying is that these design criteri...}\\\hline

\RaggedRight{Washington, Jan. 27. —(P)—Im- mediate removal of John F. J. Her- bert, as prohibition administrator for Montana and Idaho, was de- manded in the senate today by Sen- ators Borah, Idaho, and Wheéeler, Montana, on the ground of charges placed before them by department of justice investigators. Wheeler accompanied his demand (Continued on Page 2)}&
\RaggedRight{| Washington, Jan. 27 1 AP).—Immiedl- aie mmoval of John F. Herbert as pro- | hibition administrator for Montana and ‘Idaho was demanded m the Seuate to- ‘day by Senators Borah. idaho, and Waeeler, Montana. on the ground of charges placed before them by Depart- meat of Justice investigators. Wheeler accompanied his demand nith a declaration that prohibition en- foreemen: had brukea down. He blamed the “politicians” and called upon the Law Enforcement Commussion to sum- mon members of the Republican Na- tona...}\\\hline

\RaggedRight{By RAYMOND CLAPPEA (Dnited Presa Stal Correspandoayy London, Jai, 38—(UP}—-The Am ‘erlcnn delegation to the navat confer ence today won {ls demand for pre- sentation: of the cnse of suxiliary warships limitation first at tho noxt plenary session Thuvaday, ‘Tho chlet delegates, mectittg at St. James palace, also decided that tho plenary sesslon would discuss the Main conference questions in alpha betical order of ihe countriea pro- posing. Press ta be Admitted The American delegation woo a second victory whe...}&
\RaggedRight{London, Jan. 24, W.P—The Amer- jean delegation fo the naval cen- ference teday won its demand for presentation of the case of auxil- jary warships linsitation flrst at the next pletrary session ‘Vhursday, The chief delegates, meeting at Si, James Pelace, also decided that the plenary session would discuss the main confeyence questions in alphabetical order af the cauntries proposing. The American delegation won a second victory when it was decided to udmil certain representatives of the press at fie plenary...}\\\hline
\end{tabularx}

\caption{Example false negatives for \rs on the NEWS-COPY dataset (pairs of texts not detected as near-duplicates by \rs but labeled as near-duplicates in the original dataset). Examples are randomly selected and truncated at 512 characters for display.}%
\label{fig:app:fns}%
\end{table}

\begin{table}[h]
\centering
\begin{tabularx}{\textwidth}{|X|X|}\hline
Text 1 & Text 2 \\ \hline

\RaggedRight{BOZEMAN, Mont. (AP) — Chet Huntley, whose resonant voice and rough-hewn face be- came familiar to millions on the nightly television news, died Wednesday in his mountain resort home. He was 62. He underwent surgery for lung cancer in January but had remained activesuntil recent weeks. He died at 2:20 a.m, according to his widow, Tippy Hunt.cy. Huntiey was teamed for 14 years with David Brinkley on NBC's Huntley-Brinkley Re- port. He quit in 1970 and re- turned to his native Montana to develop the \$20-millio...}&
\RaggedRight{BOZEMAN, Mont. (AP) - Chet Huntley, whose resonant voice and rough-hewn face became familiar to millions on the nightly television news, died Wednesday in his mountain resort home. He was 62. He underwent surgery for lung cancer in January but had remained active until recent weeks. He died at 2:20 a.m., according to his widow, Tippy Huntley. Huntley was teamed for 14 years with David Brinkley on NBC’s Huntley- Brinkley Report. He quit in 1970 and returned to his native Montana to develop the \$20 million Bi...}\\\hline

\RaggedRight{By THE ASSOCIATED PRESS Some Americans are paying up to 50 per cent more per month for electricity this year than they did last, an Associ- ated Press survey shows. Consumers are beginning to organize to fight the rate hikes. A spot check of monthly elec- tric bills this year and last showed that most increases have been about \$1 or \$2, gen- erally about 10 per cent, with the highest reported boost com- ing in Jacksonville, Fia., where the average tab went from \$17.90 last year to \$27.70 this year. Utility...}&
\RaggedRight{By Louise Cook Acenciaiod Prece Writer Same Americans are paying up io 20 per cent more per month far electricity this year ihan they did last, an -Associ- Press survey shows. onsumers are beginning to ze to fight the rate hikes, A spot check of monthly elec- tre hills this year and Jast showed that most increases ve been about \$1 or \$2, gen- erally about 10 per cent, with the highest reported boost com- ing in Jacksonville, Fla., where the average tab went from \$17.90 last year to \$27.70 this year...}\\\hline

\RaggedRight{BOZEMAN, Mont. (AP) — Vice President Gerald R. Ford says the world will miss the “‘unique abilities” of former television news anchorman Chet Huntley. Huntley, 62, died at his home Wednesday after a long bout with lung cancer. Family ‘spokesmen said a memorial service would be conducted for Huntley Sunday at the Big Sky of Montana’ resort and recreation area south of Bozeman. Huntley was chairman of the Big Sky board of directors. Another memorial service is scheduled Tuesday in the New York studios of the...}&
\RaggedRight{BOZEMAN, Mont. (AP) — Vice President Gerald R. Ford says the world will miss the “unique abilities’’ of former television news anchorman Chet Huntley. Huntley, 62, died at his home Wednesday after a long bout with lung cancer. Family spokesmen said a me- morial service would be con- ducted for Huntley Sunday at the Big- Sky of Montana resort and recreation area south of Bozeman. Huntley was chair- man of the Big Sky board of directors. Another memorial service is scheduled Tuesday in the New York studios of...}\\\hline

\RaggedRight{WASHINGTON (AP) — The House has passed legislation raising the minimum wage from \$1.60 an hour to \$2 this year for most workers covered and to \$2.30 for all by 1978. The bill, approved Wednesday 375 to 37, also would increase by 7 million to 56.5 million the number of workers covered by the minimum wage laws. The bill is a modified version of one President Nixon vetoed last year. However, he is expected to sign this one if it is finally approved after ad- justment with a similar Senate passed measure, altho...}&
\RaggedRight{\_ WASHINGTON (AP) — The House has passed legislation raising the minimum wage from \$1.60 an hour to \$2 this year for most workers covered and to \$2.30 for all by 1978. The bill, approved Wednes- day 375 to 37, also would in- crease by 7 million to 56.5 mil- lion the number of workers cov- ered by the minimum wage laws. The bill is a modified version of one President Nixon vetoed last year. However, he is ex- ted to sign this one if it is inally approved after adjust- ment with a similar Senate- passed measu...}\\\hline

\end{tabularx}

\caption{Example false positives for \rs on the NEWS-COPY dataset (pairs of texts detected as near-duplicates by \rs but not labeled as near-duplicates in the original dataset). Examples are randomly selected and truncated at 512 characters for display.}%
\label{fig:app:fps}%
\end{table}